\documentclass[UTF8,a4paper,11pt]{article} % Changed from ctexart to article

\usepackage[margin=2.5cm]{geometry}
\usepackage{amsmath,amssymb,amsthm}
\usepackage{graphicx}
\usepackage{booktabs}
\usepackage{array}
\usepackage{multirow}
\usepackage{float}
\usepackage{subcaption}
\usepackage{algorithm}
\usepackage{algpseudocode}
\usepackage{enumitem}
\usepackage[hidelinks]{hyperref}
\graphicspath{{img/}}

\newcommand{\R}{\mathbb{R}}
\newcommand{\softmax}{\mathrm{softmax}}

\newcommand{\cC}{\mathcal{C}}
\newcommand{\cD}{\mathcal{D}}
\newcommand{\cU}{\mathcal{U}}

\newcommand{\cF}{\mathcal{F}}

\title{\bfseries ProxyFormer: A Dual-Stream Proxy Architecture for Ultra-Long Context and High-Resolution Generation}
\author{Zhongpan Tang\\
\small Independent Researcher\\
\small \texttt{tangzhongp@qq.com}}
\date{}

\begin{document}

\maketitle

\begin{abstract}
The quadratic growth of attention computation and key-value (KV) cache with respect to sequence length is a central bottleneck for ultra-long-context language models and high-resolution generative models. We propose \textbf{ProxyFormer}, a general dual-stream architecture built upon \emph{proxy tokens}. In each layer, fine-grained local features are compressed bottom-up into a small set of proxy states; expensive global interactions are performed only in the compressed proxy space; the globally contextualized proxies are then decompressed and injected top-down back into the local stream. Because the local stream persists across layers, fine-grained information that is not captured by one compression step remains accessible for later refinement, alleviating the irreversible information loss of conventional one-shot compression. We further introduce factorized multi-level compression/decompression, layer-wise dynamic compression ratios, asymmetric dual embeddings, and a proxy-only KV-cache inference scheme. On a 16\,GB GPU with batch size 1, a standard decoder-only model can train sequences of only about 20K tokens, whereas ProxyFormer with a compression ratio of 64 extends the trainable sequence length to about 0.7M. A model trained with a 64K window retains 92\%--95\% retrieval accuracy on a multi-needle retrieval task with 1,048,576 tokens, and a model trained with an 8K window exceeds 94\% accuracy when extrapolated to 256K tokens. Preliminary image-generation experiments demonstrate the feasibility of ProxyFormer for both pixel-space and latent-space flow matching.
\end{abstract}

\noindent\textbf{Keywords:} long-sequence modeling; efficient attention; proxy token; KV-cache compression; length extrapolation; flow matching

\section{Introduction}

The attention mechanism \cite{vaswani_attention_2023} has established the Transformer as the foundational architecture for language modeling, long-context reasoning, and visual generation. For a sequence of length $L$ and feature dimension $d$, the computational complexity of standard self-attention is $O(L^2d)$; in autoregressive decoding, the key and value of each historical token must also be persistently cached, resulting in a single-layer KV cache size of $O(Ld)$. As context lengths reach the hundreds of thousands or millions, attention computation and KV caching often far exceed the model weights themselves, becoming the primary factors limiting inference throughput and training sequence length \cite{dao_flashattention_2022,dao_flashattention-2_2023}.

To address this problem, existing work has largely followed three routes: first, designing linear, sparse, or fixed-pattern attention approximations \cite{wang_linformer_2020,beltagy_longformer_2020,zaheer_big_2021}; second, replacing attention with state space models or recurrent forms \cite{gu_mamba_2024,dao_transformers_2024}; third, pruning, quantizing, or evicting the KV cache \cite{zhang_h_2o_2023,xiao_efficient_2024,munkhdalai_leave_2024}. While effective in their respective scenarios, these methods generally face a structural contradiction: \emph{if fine-grained information is discarded during compression or approximation, the capability for precise long-range retrieval degrades; if all fine-grained information is retained, the computational and memory advantages are limited}. Single-shot, irreversible sequence compression is particularly prone to the "lost in the middle" phenomenon and degradation in Needle-In-A-Haystack (NIAH) performance.

This paper introduces ProxyFormer, driven by the motivation to decouple \emph{information fidelity} and \emph{global interaction cost}. ProxyFormer does not require all tokens to participate directly in global attention, but instead organizes the input into two parallel feature streams:
\begin{enumerate}[leftmargin=2em]
    \item \textbf{Fine/local stream}: maintains fine-grained features that correspond one-to-one with input elements, responsible for preserving local details;
    \item \textbf{Proxy stream}: a small number of proxy vectors obtained by compressing local blocks, responsible for completing global interactions at an extremely low cost.
\end{enumerate}
Each layer executes a closed loop of "bottom-up compression $\rightarrow$ proxy space interaction $\rightarrow$ top-down decompression and injection". Because the local stream maintains residual connections across layers, details not absorbed by a compression step in one layer can still be passed to deeper layers and re-utilized in subsequent compressions; meanwhile, the global information carried by the proxy stream is reversely injected into the local stream through decompression and mechanisms such as gating, modulation, or cross-attention. This mechanism allows the model, at a compression ratio of $P$, to reduce the sequence length for proxy space attention to $L/P$ while preserving the information integrity of the local stream.

The main contributions of this paper are as follows:
\begin{itemize}[leftmargin=2em]
    \item We propose the general ProxyFormer architecture, uniformly supporting 1D sequences, 2D images, 3D point clouds/voxels/videos, and higher-dimensional tensors, covering tasks such as autoregressive language modeling, flow matching image generation, and diffusion conditional generation;
    \item We provide an implementation-friendly architectural instance: compression and decompression are composed of standard operators such as reshape, linear projection, and convolution; proxy interactions can directly reuse existing attention or SSM modules without the need for sparse indexers, block-sparse attention kernels, or custom CUDA operators;
    \item We introduce a bidirectional communication mechanism between the local and proxy streams, as well as factorized multi-level compression/decompression and layer-wise dynamic compression ratios, which control the parameter size of the compression modules while maintaining high compression ratios;
    \item We propose a proxy-level KV cache and a streaming block-triggered update strategy, allowing autoregressive inference to only maintain historical proxy states of length $L/P$, while original historical local features can be released or offloaded;
    \item We systematically validate the architecture through preliminary experiments on language model perplexity, multi-needle NIAH, memory/speed benchmarks, and image generation tasks, providing quantitative comparisons with standard decoder-only baselines.
\end{itemize}

\section{Related Work}

\subsection{Efficient Attention and Long-Context Modeling}
Linear attention and kernel methods approximate attention in a low-rank or feature-mapped form, reducing complexity to $O(L)$ \cite{katharopoulos_transformers_2020}. Linformer demonstrates that the self-attention matrix can be low-rank approximated and uses a fixed number of learnable projections to reduce complexity to $O(L)$ \cite{wang_linformer_2020}. Longformer, BigBird, and others implement sparse attention by combining sliding windows, dilated windows, and global tokens \cite{beltagy_longformer_2020,zaheer_big_2021}. The FlashAttention series accelerates exact attention from an I/O perspective but does not alter its quadratic complexity \cite{dao_flashattention_2022,dao_flashattention-2_2023}. State space models formulate sequence modeling as a linear time-varying system, with Mamba and its extensions demonstrating competitive quality and efficiency on long sequences \cite{gu_mamba_2024,dao_transformers_2024}. Unlike the aforementioned methods, ProxyFormer does not require changing the mathematical form of the global interaction operator; its interaction operator can be attention, SSM, RNN, or an identity mapping. The compression benefits primarily stem from the structural property that interactions occur in the extremely short proxy space.

\subsection{KV Cache Compression and Token Selection}
To reduce KV cache usage in autoregressive inference, StreamingLLM retains "attention sinks" and recent tokens \cite{xiao_efficient_2024}; H$_2$O evicts low-scoring tokens based on attention scores \cite{zhang_h_2o_2023}; Infini-attention compresses long-term memory into a fixed number of reusable states \cite{munkhdalai_leave_2024}. Landmark Attention and similar memory token methods compress historical blocks into summary tokens and cache them \cite{mohtashami_landmark_2023}. A commonality among these methods is that compression or eviction typically occurs after attention, and original tokens are unrecoverable once discarded. ProxyFormer, on the other hand, places compression \emph{before} global interaction and treats historical proxy states as a trainable cross-layer stream; more importantly, the local stream remains present during training, enabling the compressor to receive fine-grained gradient signals from deeper layers.

\subsection{Efficient Representations in Visual Generation}
Diffusion models \cite{ho_denoising_2020} and flow matching \cite{lipman_flow_2023} have become mainstream paradigms for high-quality image generation. To reduce the computational cost of high-resolution generation, common practices include generating in a VAE latent space first \cite{rombach_high-resolution_2022,kingma_auto-encoding_2022} or applying local window attention to feature maps. ProxyFormer provides a unified patch-to-proxy compression pathway: for pixel-space (JIT) or latent-space (JLT) feature maps, proxy vectors are extracted based on 2D patches, and cross-modal conditions only perform cross-attention with the proxy vectors. This significantly reduces the cost of condition interaction and global attention while maintaining the local pixel/latent feature stream.

\subsection{Position of this Work}
Compared to fixed-length compressed attention works such as TLinformer and TConstFormer \cite{tang_rethinking_2025,tang_tlinformer_2025}, ProxyFormer's compressed length can dynamically vary with the input length and network layers, and it features symmetric decompression and top-down injection pathways. Compared to the study of near-lossless compression on fixed-shape features in "Compression is Routing" \cite{tang_compression_2025}, this paper systemizes that concept into a trainable, stackable dual-stream architecture applicable to autoregressive decoding and visual generation.

\section{ProxyFormer Methodology}

\subsection{Problem Definition and Overall Architecture}

\begin{figure}[H]
\centering
\includegraphics[width=0.86\textwidth]{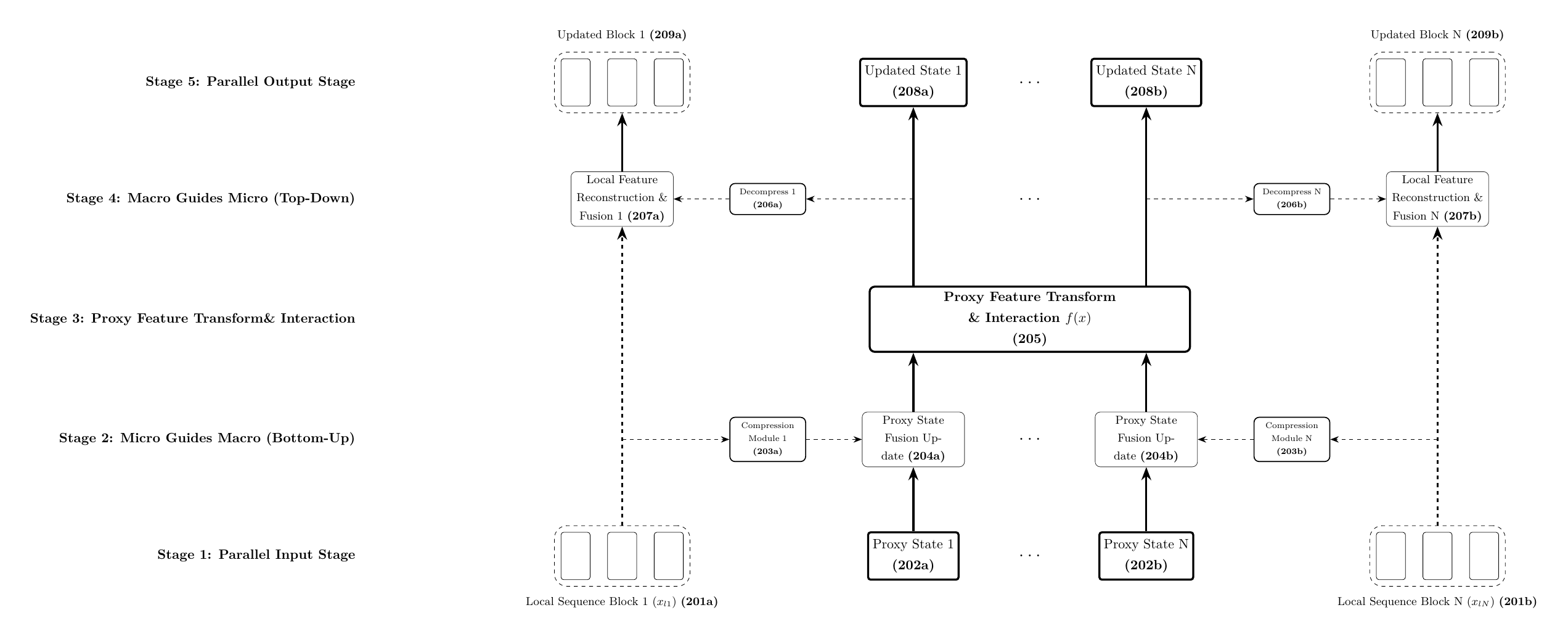}
\caption{Local stream and proxy stream dual-stream architecture. Fine-grained local features update proxy states through "micro guiding macro"; after proxy states complete global interaction, they are injected into the local stream through "macro guiding micro".}
\label{fig:dual}
\end{figure}

Formally, let the input features be a general tensor $\mathcal X^{(0)}\in\R^{N_1\times\cdots\times N_D\times d_{\mathrm{in}}}$, where $D$ is the number of spatial/temporal/topological dimensions, $N_1,\dots,N_D$ are the sizes of each dimension, and $d_{\mathrm{in}}$ is the number of channels. For convenience, we flatten it into a feature sequence $X^{(0)}\in\R^{L\times d_{\mathrm{in}}}$ of length $L=\prod_{j=1}^{D}N_j$; for text, $D=1$; for images, $D=2$; for 3D point clouds/voxels or spatiotemporal video, $D=3$ can be used; higher-dimensional data can be recursively grouped or flattened. A standard Transformer layer can be written as
\begin{equation}
A^{(l)}=\softmax\!\left(\frac{QK^{\top}}{\sqrt{d_{\mathrm{head}}}}\right)V,\qquad
X^{(l)}=X^{(l-1)}+\mathrm{FFN}^{(l)}(A^{(l)}),
\label{eq:full-attn}
\end{equation}
where $Q,K,V$ are obtained by projecting $X^{(l-1)}$. The computation and activation cache of Equation \eqref{eq:full-attn} are both $O(L^2)$. ProxyFormer rewrites each layer into the following five steps:

\begin{equation}
\{X_1^{(l-1)},\dots,X_m^{(l-1)}\}=\mathrm{Partition}\!\left(X^{(l-1)}; P^{(l)}\right),
\label{eq:partition}
\end{equation}
\begin{equation}
c_i^{(l)}=\cC_i^{(l)}\!\left(X_i^{(l-1)}\right),\qquad
H_i^{(l)}=\cU_i^{(l)}\!\left(H_i^{(l-1)},c_i^{(l)}\right),
\label{eq:bottom-up}
\end{equation}
\begin{equation}
\bar H^{(l)}=\cF^{(l)}\!\left(H^{(l)}\right),
\label{eq:proxy-interaction}
\end{equation}
\begin{equation}
\Delta_i^{(l)}=\cD_i^{(l)}\!\left(\bar H_i^{(l)}\right),
\label{eq:top-down-decomp}
\end{equation}
\begin{equation}
X^{(l)}=\Phi^{(l)}\!\left(X^{(l-1)},\Delta_1^{(l)},\dots,\Delta_m^{(l)}\right),
\label{eq:injection}
\end{equation}
where $P^{(l)}$ is the block size for the $l$-th layer, $m=\lceil L/P^{(l)}\rceil$; $X_i\in\R^{P^{(l)}\times d_{\mathrm{fine}}}$ is the $i$-th local block; $H_i\in\R^{d_{\mathrm{proxy}}}$ is the corresponding proxy vector; $\cC,\cD$ are the compression and decompression modules respectively; $\cU$ is the proxy state update; $\cF$ is the proxy space interaction; and $\Phi$ is the local feature injection and fusion. Algorithm \ref{alg:layer} summarizes the forward process of a ProxyFormer layer, and Figure \ref{fig:dual} depicts the cross-layer dual-stream information pathway.

\begin{algorithm}[H]
\caption{Forward Process of a Single ProxyFormer Layer}
\label{alg:layer}
\begin{algorithmic}[1]
\Require Local stream $X^{(l-1)}\in\R^{L\times d_{\mathrm{fine}}}$, proxy stream $H^{(l-1)}\in\R^{m\times d_{\mathrm{proxy}}}$, block size $P^{(l)}$
\Ensure Updated local stream $X^{(l)}$ and proxy stream $H^{(l)}$
\State $\{X_1,\dots,X_m\}\gets \mathrm{Partition}(X^{(l-1)};P^{(l)})$
\For{$i=1,\dots,m$}
    \State $c_i\gets \cC_i(X_i)$
    \State $H_i\gets \cU_i(H_i,c_i)$
\EndFor
\State $\bar H\gets \cF(H)$
\For{$i=1,\dots,m$}
    \State $\Delta_i\gets \cD_i(\bar H_i)$
\EndFor
\State $X^{(l)}\gets \Phi(X^{(l-1)},\Delta_1,\dots,\Delta_m)$
\State \Return $X^{(l)},H^{(l)}$
\end{algorithmic}
\end{algorithm}

\subsection{Bottom-Up Compression}
\paragraph{Partitioning.}
For 1D sequences, we adopt contiguous, non-overlapping block partitioning, though sliding overlapping windows are also supported; for 2D images, partitioning degenerates into spatial patches; for 3D voxels, point clouds, or spatiotemporal videos, one can use 3D voxel blocks, spatial neighborhood grouping, voxelization followed by partitioning, or dynamic clustering based on semantics/distance; for higher-dimensional tensors, they can be recursively flattened and partitioned. Let the $i$-th flattened block be $X_i\in\R^{P\times d_{\mathrm{fine}}}$. In our experiments, we use fixed boundary partitioning; more general dynamic routing/clustering grouping is a generalization of Equation \eqref{eq:partition}.

\paragraph{Proxy Extraction.}
The compression module $\cC_i$ maps $X_i$ to a low-dimensional vector $c_i$. The instance used in our experiments is "reshape--linear projection":
\begin{equation}
c_i=\mathrm{Proj}\!\left(\mathrm{Flatten}\!\left(X_i\right)\right)\in\R^{d_{\mathrm{proxy}}},
\label{eq:compress-linear}
\end{equation}
For 2D patches, spatial flattening and 2D convolution/linear projection can be used; for 3D voxel/video blocks, 3D convolution or corresponding high-dimensional unfolding can be used. These operators are standard deep learning primitives and do not require custom sparse indexers. The proxy state update employs a residual form:
\begin{equation}
H_i^{(l)}=H_i^{(l-1)}+c_i^{(l)},
\label{eq:proxy-res}
\end{equation}
which can also be replaced by gated updates, concatenated projection, or direct substitution. When a layer has no cross-layer proxy state input, $H_i^{(l-1)}=0$, and Equation \eqref{eq:proxy-res} degenerates to direct compression.

\paragraph{Asymmetric Dual Embeddings.}
In autoregressive language models, the historical region and the generation region bear different cost structures. ProxyFormer uses a low-dimensional embedding $E_h:\mathcal V\to\R^{d_h}$ for the historical region and a full-dimensional embedding $E_x:\mathcal V\to\R^{d_{\mathrm{model}}}$ for the generation region, with their parameters independent and dimensions decoupled. In our experiments, $d_h=64$ and $d_{\mathrm{model}}=512$, meaning the historical sequence is compressed by a factor of 8 in channel memory at the very first step entering the network. This design is compatible with the dimensional decoupling of the local and proxy streams, and can also be replaced by shared embeddings with projection.

\subsection{Global Interaction in Proxy Space}
The proxy interaction function $\cF^{(l)}$ is a pluggable operator in the architecture. Let $\bar H\in\R^{m\times d_{\mathrm{proxy}}}$, its optional forms include:
\begin{itemize}[leftmargin=2em]
    \item \textbf{Identity/Pass-through}: $\bar H=H$, suitable for completely independent block evolution;
    \item \textbf{Single-Proxy MLP}: applying a shared feed-forward network independently to each proxy vector;
    \item \textbf{Causal/Bidirectional Attention}:
    \begin{equation}
    \bar H=\softmax\!\left(\frac{Q_pK_p^{\top}}{\sqrt{d_{\mathrm{head}}}}+M\right)V_p,
    \label{eq:proxy-attn}
    \end{equation}
    where $M$ is a causal mask (for autoregressive tasks) or a zero matrix (for bidirectional tasks);
    \item \textbf{State Space or Recurrent Operators}: feeding the proxy sequence into an SSM/RNN;
    \item \textbf{Cross-Modal Cross-Interaction}: using proxy vectors as queries to perform cross-attention with text, time steps, or control conditions.
\end{itemize}
Since the sequence length in Equation \eqref{eq:proxy-attn} is $m\approx L/P$, the cost of proxy space attention is approximately $P^2$ times lower than full attention. This property is independent of any specific low-rank or kernel approximation assumptions.

\subsection{Top-Down Decompression and Injection}
For each block, the interacted proxy vector is decompressed back to the local space:
\begin{equation}
\Delta_i=\mathrm{Unflatten}\!\left(\mathrm{Proj}^{\uparrow}\!\left(\bar H_i\right)\right)\in\R^{P\times d_{\mathrm{fine}}}.
\label{eq:decompress-linear}
\end{equation}
The local fusion $\Phi$ can choose from the following mechanisms:
\begin{align}
\text{Residual Addition:}\quad & X_i\leftarrow X_i+\Delta_i,\label{eq:inj-add}\\
\text{Gated Fusion:}\quad & X_i\leftarrow X_i+\sigma\!\left(g(X_i,\Delta_i)\right)\odot\Delta_i,\label{eq:inj-gate}\\
\text{Affine Modulation:}\quad & X_i\leftarrow \left(1+s(\Delta_i)\right)\odot \mathrm{Norm}(X_i)+t(\Delta_i),\label{eq:inj-affine}\\
\text{Cross-Attention:}\quad & X_i\leftarrow X_i+\mathrm{Attn}\!\left(Q=X_i,\,K=\bar H_i,\,V=\bar H_i\right).\label{eq:inj-cross}
\end{align}
Subsequently, the local stream can continue to be refined by a local Transformer or convolution. Our language model implementation uses gated cross-attention (a gated variant of Equation \eqref{eq:inj-cross}) to inject proxy history into the generation region; details are in Section \ref{sec:llm}.

The key design is that \emph{compression is not the destruction of the local stream, but the summarization of it}. Even if the compressor at layer $l$ misses a detail, that detail is preserved in the residual stream of $X^{(l)}$ and can participate in compression again at layer $l+1$. Therefore, the compression bias of ProxyFormer can be corrected through cross-layer iteration, which is fundamentally different from methods that discard original tokens after a single compression.

\subsection{Multi-Level Factorized Compression and Decompression}

\begin{figure}[H]
\centering
\includegraphics[width=0.92\textwidth]{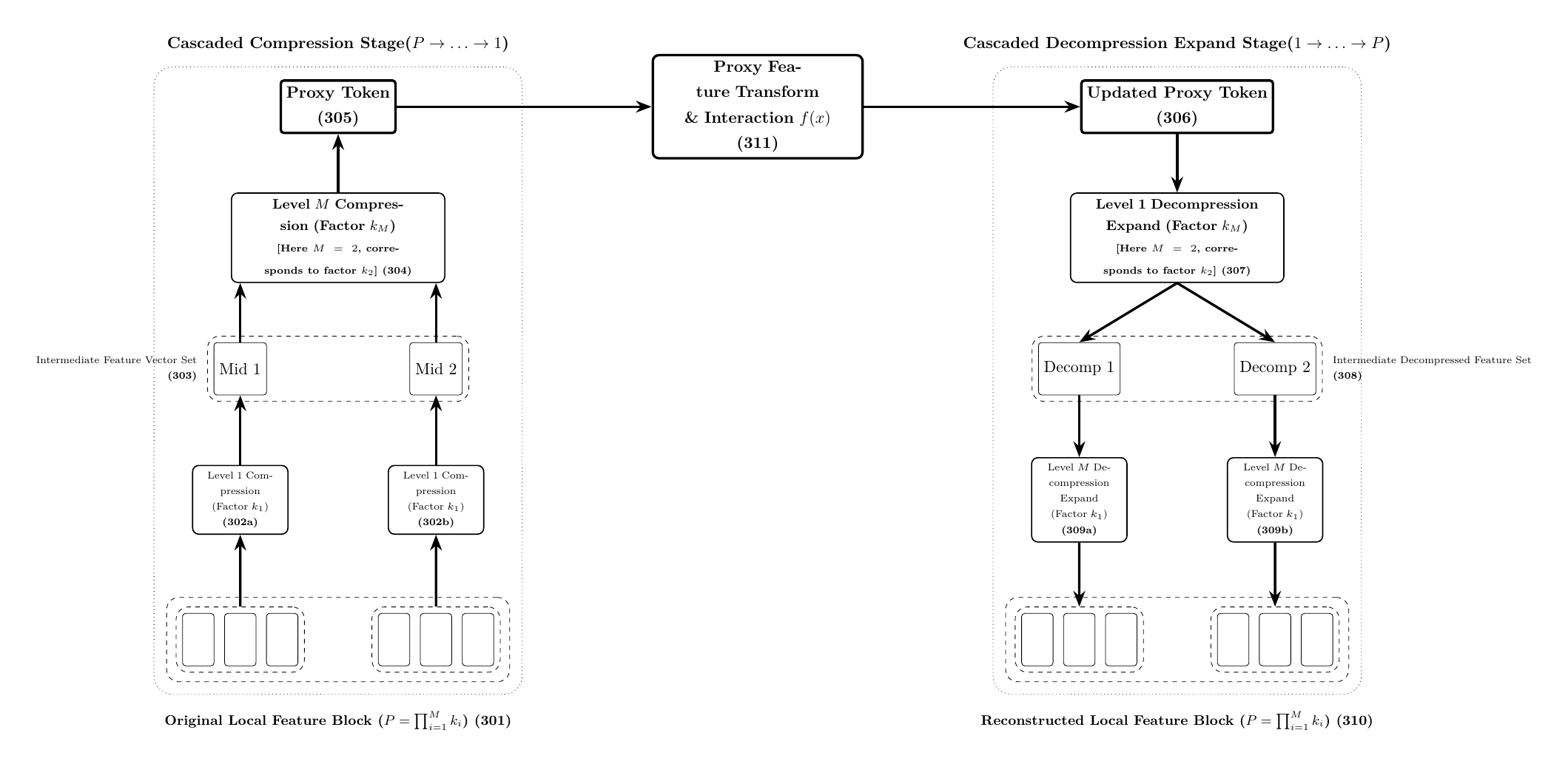}
\caption{Multi-level cascaded factorized compression and symmetric decompression. $P=\prod_{j=1}^M k_j$, and intermediate features exist only on shorter sequences with narrower channels.}
\label{fig:factorization}
\end{figure}

When the compression ratio $P$ is large, the parameters and computation of a single linear projection are both $O(Pd_{\mathrm{fine}}d_{\mathrm{proxy}})$, which is prone to overfitting or becoming untrainable. ProxyFormer factorizes $P$ into $M$ integer factors:
\begin{equation}
P=\prod_{j=1}^{M}k_j,\qquad k_j\ge 1,
\label{eq:factor}
\end{equation}
and constructs a cascaded compression:
\begin{equation}
\cC=\cC_M\circ\cC_{M-1}\circ\cdots\circ\cC_1,\qquad
\cC_j:\R^{K_{j-1}\times d_{j-1}}\to\R^{K_j\times d_j},
\label{eq:factor-cascade}
\end{equation}
where $K_j=K_{j-1}/k_j$, $K_0=P$, and $K_M=1$. A typical implementation is "reshape $\to$ project": first group $K_{j-1}$ elements into groups of $k_j$ and concatenate along the channel dimension to obtain $\R^{K_j\times(k_jd_{j-1})}$, then linearly project to $d_j$. The decompression side executes reverse reshape and up-projection in the opposite order:
\begin{equation}
\cD=\cD_1\circ\cD_2\circ\cdots\circ\cD_M,\qquad
\cD_j:\R^{K_j\times d_j}\to\R^{K_{j-1}\times d_{j-1}}.
\label{eq:factor-dec}
\end{equation}
Figure \ref{fig:factorization} illustrates a two-level factorization. If the parameter size of each sub-projection is approximately $k_jd_{j-1}d_j$, the total parameter size on the compression side is $\sum_{j}O(k_jd_{j-1}d_j)$, whereas a single mapping is $O(Pd_{\mathrm{fine}}d_{\mathrm{proxy}})$. For example, when $P=64$, $M=3$, and $k_j=4$, the reshape-projection pathway significantly reduces the dimensionality of single-layer matrices.

Conv/ConvTranspose downsampling and upsampling in visual generation can be viewed as 2D special cases of Equations \eqref{eq:factor-cascade}--\eqref{eq:factor-dec}; 3D convolutions and their transposes can similarly be viewed as corresponding implementations for 3D data.

\subsection{Layer-wise Dynamic Compression Ratios and Memory Hierarchy}
ProxyFormer allows different layers to use different $P^{(l)}$. When the compression ratio is inconsistent between adjacent layers, the proxy state $H$ is re-initialized and re-extracted from the local stream of the current layer; this is equivalent to "re-partitioning" across different scales of abstraction. Furthermore, different compression ratios can be configured for historical blocks at different temporal distances: distant history uses coarser-grained proxies, while recent history retains more fine-grained tokens. This horizontal/vertical dynamic configuration requires altering attention masks and cache layouts in standard Transformers, whereas in ProxyFormer it only involves hyperparameters for partitioning and compression modules.

\subsection{Language Model Instantiation}
\label{sec:llm}

\begin{figure}[H]
\centering
\includegraphics[width=0.94\textwidth]{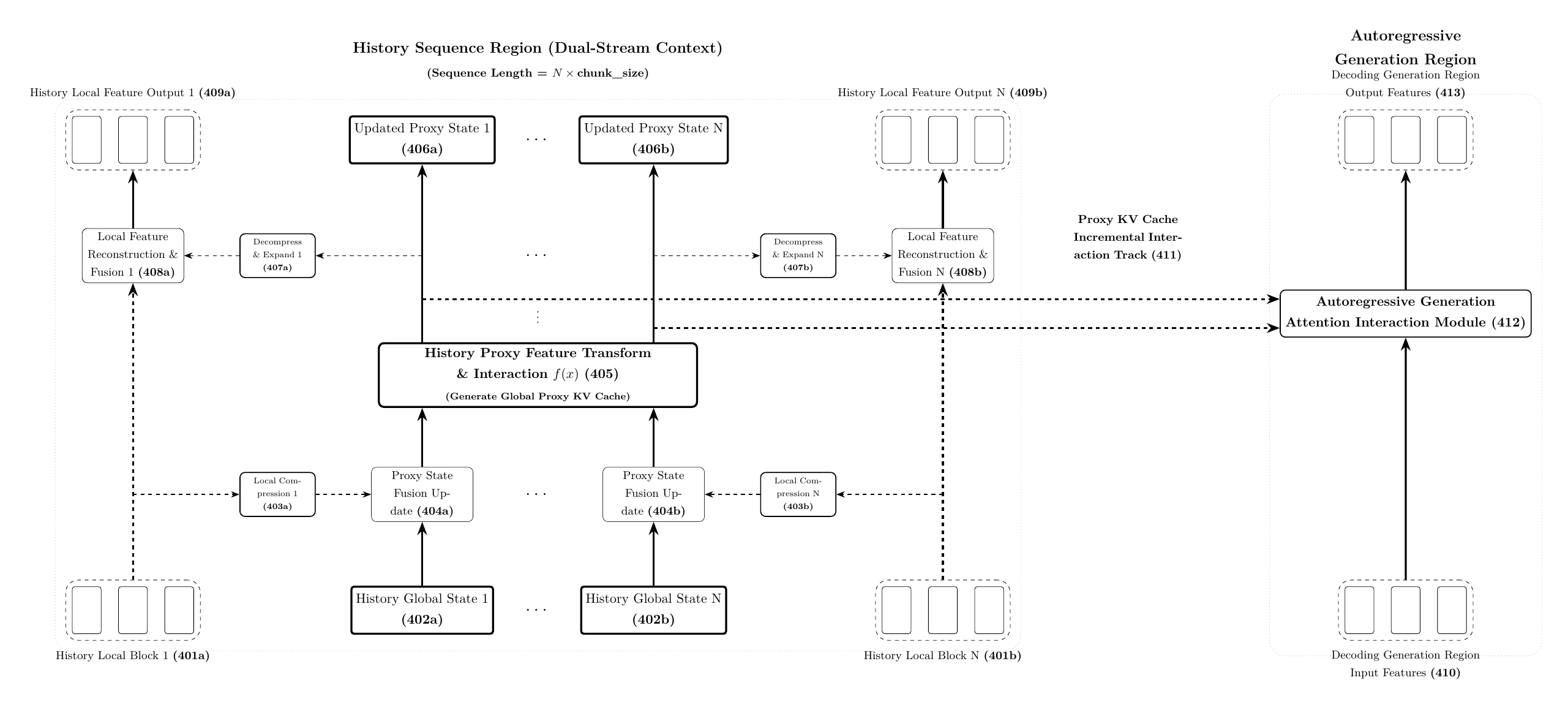}
\caption{Language model based on proxy features. Long history is compressed into an extremely small proxy KV cache; during decoding, the generation region only performs gated cross-attention with the proxy cache.}
\label{fig:llm}
\end{figure}

Figure \ref{fig:llm} illustrates the autoregressive language model based on ProxyFormer. Given a history token sequence $t_h$ and a generation region token sequence $t_x$, the model first computes
\begin{equation}
H^{(0)}_{\mathrm{fine}}=E_h(t_h)\in\R^{L_h\times d_h},\qquad
X^{(0)}=E_x(t_x)\in\R^{L_x\times d_{\mathrm{model}}}.
\end{equation}
On the historical side, $L_{\mathrm{depth}}$ ProxyFormer blocks are stacked to compress $H_{\mathrm{fine}}$ into proxy state $H$; under causal settings, the proxy vectors in $H$ depend only on the current block and past blocks, and thus can be safely cached. The generation side uses standard Transformer blocks and adds cross-attention to the historical proxies alongside self-attention:
\begin{equation}
A_x=\mathrm{SDPA}\!\left(q_x,k_x,v_x;\mathrm{causal}\right),\quad
A_h=\mathrm{SDPA}\!\left(q_x,k_h,v_h\right),
\label{eq:llm-cross}
\end{equation}
\begin{equation}
\alpha=\mathrm{MLP}_{\alpha}\!\left(\mathrm{RMSNorm}(x)\right),\qquad
A=A_x+\alpha\odot A_h,
\label{eq:llm-gate}
\end{equation}
where $\alpha$ is computed per attention head, allowing the model to adaptively decide how much information to read from the historical proxies. The output layer is a shared-weight RMSNorm + linear language modeling head; the training loss is the standard next-token cross-entropy, appended with a loss term from intermediate layer auxiliary heads.

\subsection{Proxy KV Cache and Streaming Decoding}
In a standard autoregressive Transformer, the historical region caches $K_h,V_h\in\R^{L_h\times d_{\mathrm{model}}}$ at each layer. ProxyFormer executes compression and proxy interaction on historical blocks during the prefill stage, and thereafter only caches
\begin{equation}
K_{\mathrm{proxy}},V_{\mathrm{proxy}}\in\R^{(L_h/P)\times d_{\mathrm{model}}},
\label{eq:proxy-cache}
\end{equation}
The original historical local features can be released, quantized, or offloaded. During the decoding stage, each step only computes self-attention for the current token and its cross-attention with the proxy cache; the historical proxy cache remains frozen. When the new tokens in the generation region reach the block size $P$, a local compression is triggered, appending the new block to the proxy cache and releasing the original local features of that block. Equation \eqref{eq:proxy-cache} shows that in the extreme case, the historical KV cache can be reduced to $1/P$ of the standard implementation; moreover, since proxy attention during the prefill stage is only executed over $L_h/P$ proxies, prefill latency is also substantially decreased.

\subsection{Visual Generation Instantiation}
\label{sec:vision}

\begin{figure}[H]
\centering
\includegraphics[width=0.95\textwidth]{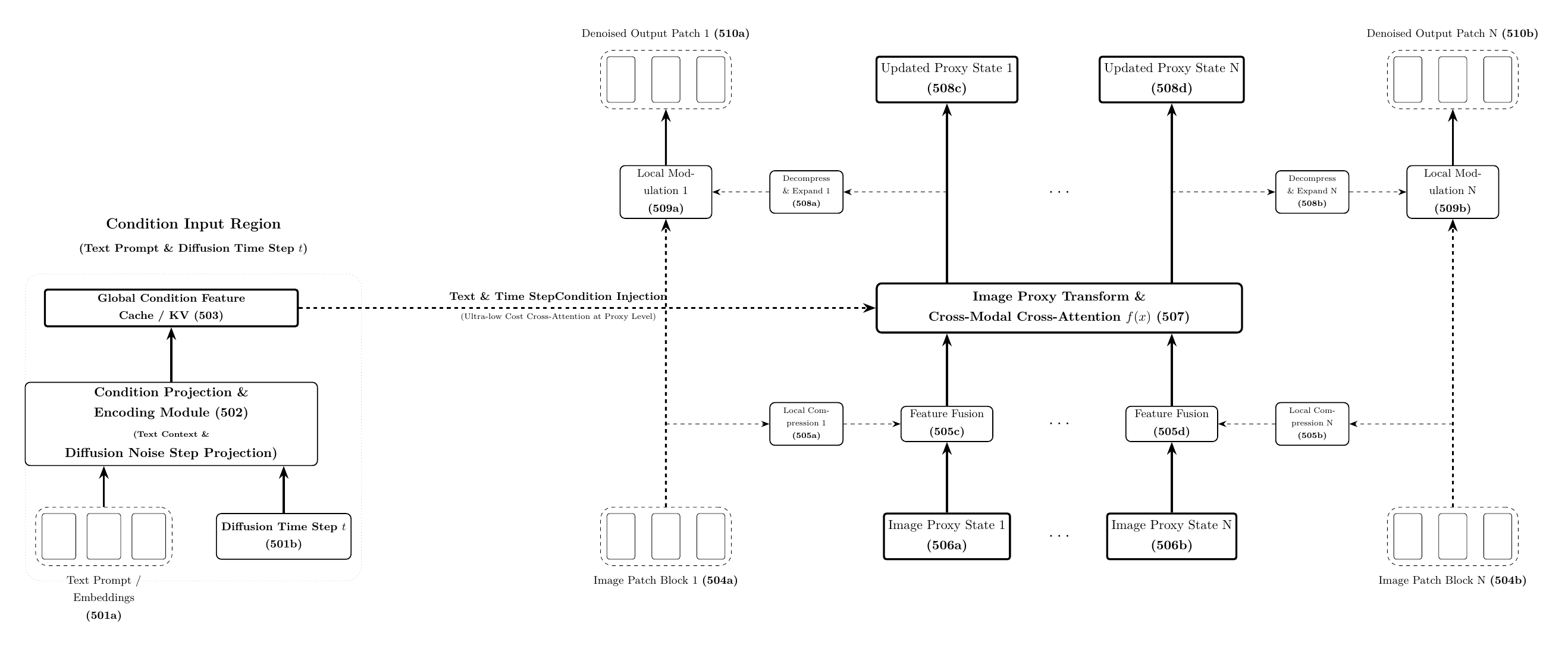}
\caption{ProxyFormer in text-to-image and diffusion conditional generation. Global conditions undergo cross-modal cross-attention only in the compressed image proxy space.}
\label{fig:visual}
\end{figure}

Figure \ref{fig:visual} depicts the ProxyFormer structure for visual diffusion/flow matching. Our image-generation experiments adopt both the JIT and JLT paradigms \cite{li_back_2026,fu_jlt_2026}. Let the image or latent tensor be $u\in\R^{C\times H\times W}$ and the patch size be $p\times p$, then the spatial compression ratio is $P=p^2$. The local stream is obtained via convolutional embedding as $X_{\mathrm{fine}}\in\R^{d_{\mathrm{fine}}\times(H/p)\times(W/p)}$; the proxy stream $H\in\R^{(H/p)\times(W/p)\times d_{\mathrm{proxy}}}$ is obtained by the 2D compression module. Text prompts, class labels, and time step conditions, once encoded, only perform cross-attention or condition injection with the proxy stream, before being symmetrically decompressed to reconstruct the local stream.

\section{Computational and Memory Complexity Analysis}
\label{sec:complexity}
To simplify the analysis, let $d=d_{\mathrm{proxy}}=d_{\mathrm{model}}$, and ignore the batch and attention head dimensions. Let the single-layer computation of full attention be $C_{\mathrm{full}}(L)=\Theta(L^2d)$ and the KV cache be $M_{\mathrm{full}}(L)=\Theta(L_{\mathrm{layer}}Ld)$, where $L_{\mathrm{layer}}$ is the number of layers. In ProxyFormer, the proxy attention length is $m=L/P$, and compression and decompression are block-wise linear operations. Thus, the single-layer computation is
\begin{equation}
C_{\mathrm{proxy}}(L)=\underbrace{\Theta\!\left((L/P)^2d\right)}_{\text{Proxy Attention}}+\underbrace{\Theta\!\left(L\,d_{\mathrm{fine}}\,d\right)}_{\text{Comp. + Decomp.}}+\underbrace{\Theta\!\left(L\,d_{\mathrm{fine}}\right)}_{\text{Local Stream}},
\label{eq:complexity}
\end{equation}
and the historical cache (proxy states only) is
\begin{equation}
M_{\mathrm{proxy}}(L)=\Theta\!\left(L_{\mathrm{layer}}\,\frac{L}{P}\,d\right).
\label{eq:memory}
\end{equation}
Therefore, when $L\gg P\,d_{\mathrm{fine}}$, the proxy attention term dominates, and the computation is approximately $1/P^2$ of full attention; when $P$ is fixed, the compression/decompression terms grow only linearly. The NIAH and memory experiments in this paper use $P=64$, giving a theoretical proxy attention speedup of 4096$\times$ and a theoretical historical KV cache compression of 64$\times$. Actual gains depend on the implementation, block size, and local stream maintenance cost.

Equation \eqref{eq:complexity} also reveals a "space-for-channel" reinvestment mechanism. Suppose the compression ratio is further increased from $P$ to $\alpha P$ ($\alpha>1$), reducing the proxy sequence length to $1/\alpha$ of its original length, and multiplying the proxy attention cost by $1/\alpha^2$; if simultaneously the proxy channel is increased from $d_{\mathrm{proxy}}$ to $\beta d_{\mathrm{proxy}}$, the proxy attention cost is multiplied by $\beta/\alpha^2$. Therefore, as long as $\beta\le\alpha^2$, one can further compress the sequence without increasing the total proxy attention budget, while significantly enhancing the information capacity of each proxy vector. In other words, compute saved on the sequence axis through compression can be continuously reinvested into the channel axis, and channel capacity in turn provides room for further increasing the compression ratio.

Equation \eqref{eq:complexity} also reveals the applicable boundaries of ProxyFormer: for short sequences or wide local streams where $d_{\mathrm{fine}}\approx d$, the linear compression term may exceed the quadratic attention term; thus, block compression is more suitable for long sequences and high-resolution inputs. By using asymmetric dual embeddings, $d_h=64\ll d=512$, the constant of the linear term is further suppressed.

\section{Experiments}

\subsection{Experimental Setup}
\paragraph{Language Modeling and NIAH.}
Language model experiments use the GPT-2 tokenizer \cite{radford2019language} (vocabulary size 50,257), model width $d_{\mathrm{model}}=512$, attention head dimension $d_{\mathrm{head}}=64$, totaling 10 layers. For ProxyFormer, the history embedding dimension is $d_h=64$, generation region block size $C=64$, and compression ratio per layer $P=64$; WikiText language modeling experiments use a block size of 128 and compression ratio of 32. Training employs a warmup-cosine learning rate schedule, gradient accumulation, and bf16 mixed precision. The optimizer is Mano (performing manifold normalized updates on 2D parameters and AdamW on others) \cite{gu_mano_2026}. The final NIAH model for evaluation turns off positional encoding (NoPE); the implementation also supports Rotary Position Embedding (RoPE) \cite{su_roformer_2023}. The baseline decoder-only model uses the same width, depth, and attention structure, with the history region and generation region sharing full-dimensional embeddings.

\paragraph{Image Generation.}
Image experiments employ Flow Matching. CIFAR-10 and MNIST images are resized to $32\times 32$. The JIT model is trained in pixel space with $4\times 4$ patches; the MNIST JLT first trains a ProxyFormer autoencoder to obtain a $16\times 16\times 4$ latent space, and then trains Flow Matching in the latent space. FID and Inception Score are calculated by torch-fidelity.

\subsection{Language Model Perplexity}
Table \ref{tab:ppl} presents the perplexity (PPL) results on WikiText-103-v1. All three sets of models turn off positional encoding. As shown: without history, ProxyFormer is close to the identically sized decoder-only baseline (22.10 vs 21.36); after introducing compressed history, the PPL drops to 21.01, outperforming the history-less baseline. This result indicates that proxy compression does not merely come at the expense of language modeling capability, but rather can bring gains through historical modeling.

\begin{table}[H]
\centering
\caption{Perplexity on WikiText-103-v1 (NoPE).}
\label{tab:ppl}
\begin{tabular}{lcc}
\toprule
Model & History & PPL $\downarrow$ \\
\midrule
Base (decoder-only) & None & 21.36 \\
ProxyFormer & None & 22.10 \\
ProxyFormer & Yes & \textbf{21.01} \\
\bottomrule
\end{tabular}
\end{table}

\subsection{Multi-Needle In A Haystack}
\label{sec:niah}
\paragraph{Task Construction.}
Inspired by the NIAH evaluation \cite{kamradt2023needle,nelson_needle_2024}, we randomly sample a background text of length $L$ tokens from WikiText-103; the background is evenly divided into $N=50$ segments, and a needle is inserted at a random position within each segment. The format of the needle is
\begin{quote}
\texttt{The passkey\_\{i\} is \{XXXXX\}. Remember it.}
\end{quote}
where $i$ is the needle index and \texttt{\{XXXXX\}} is a random 5-digit password. Each document contains 50 needles covering a depth of roughly $0\%$--$100\%$. We query each needle individually:
\begin{quote}
\texttt{What is the passkey\_\{i\}? The passkey\_\{i\} is}
\end{quote}
and perform greedy decoding (up to 10 new tokens). A hit is determined if the generated text contains the corresponding 5-digit password. We repeat 100 independent documents for each length, resulting in a total of $100\times 50=5,000$ retrievals per length. Background tokens are resampled when insufficient.

\paragraph{Curriculum Learning.}
Long-sequence NIAH training uses a four-stage curriculum: 256-token single-needle warmup, 256-token single-needle proxy training, 8K-token 50-needle training, and 8K/64K 50-needle training with disabled positional encoding. Each stage is manually stopped after the training loss approaches stabilizes, then proceeds to the next stage. Training a 64K-long context directly from scratch was found to be difficult to stably converge in our experiments.

\begin{figure}[H]
\centering
\begin{subfigure}[t]{0.49\textwidth}
\centering
\includegraphics[width=\textwidth]{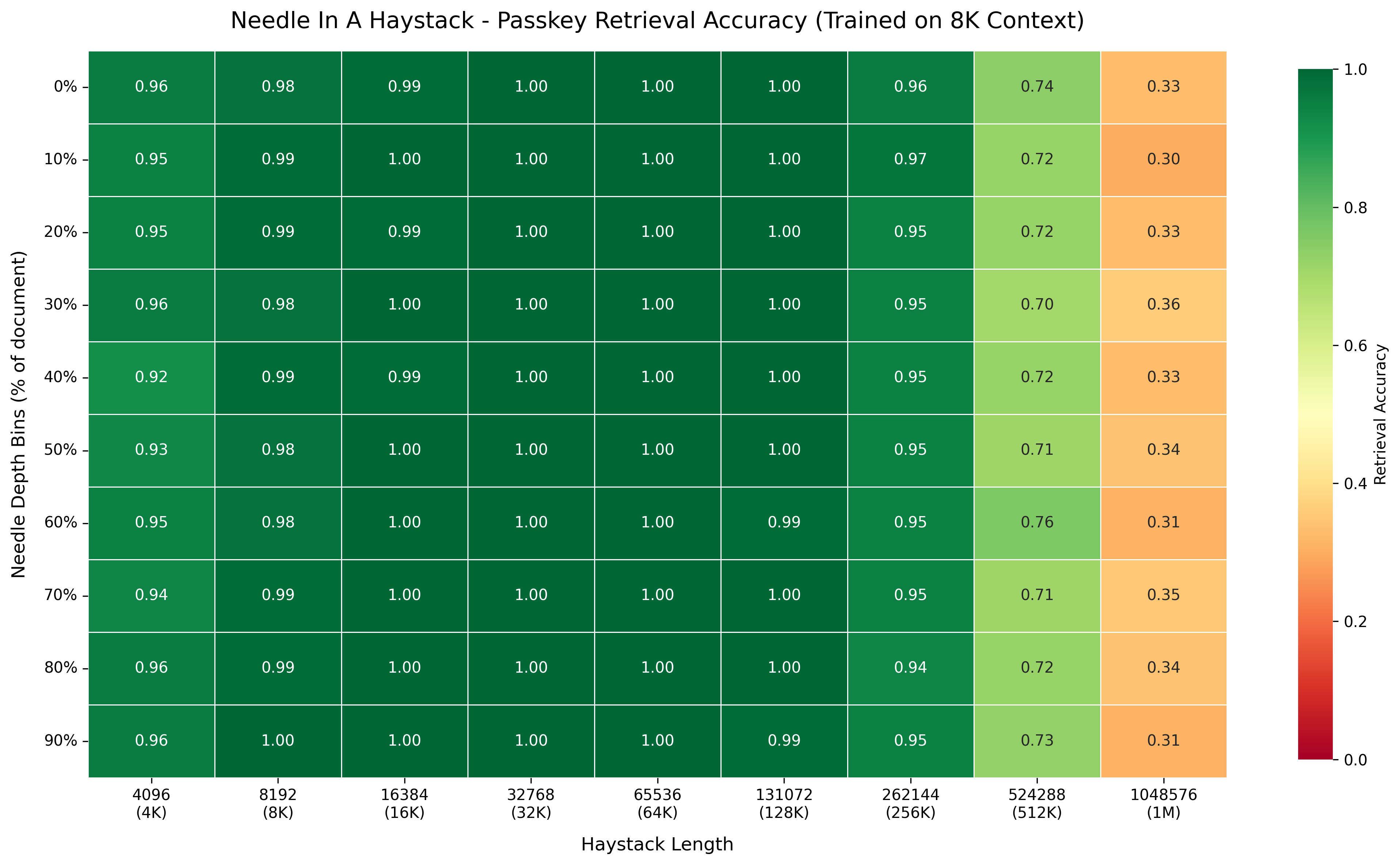}
\caption{8K Training Window}
\label{fig:niah8k}
\end{subfigure}
\hfill
\begin{subfigure}[t]{0.49\textwidth}
\centering
\includegraphics[width=\textwidth]{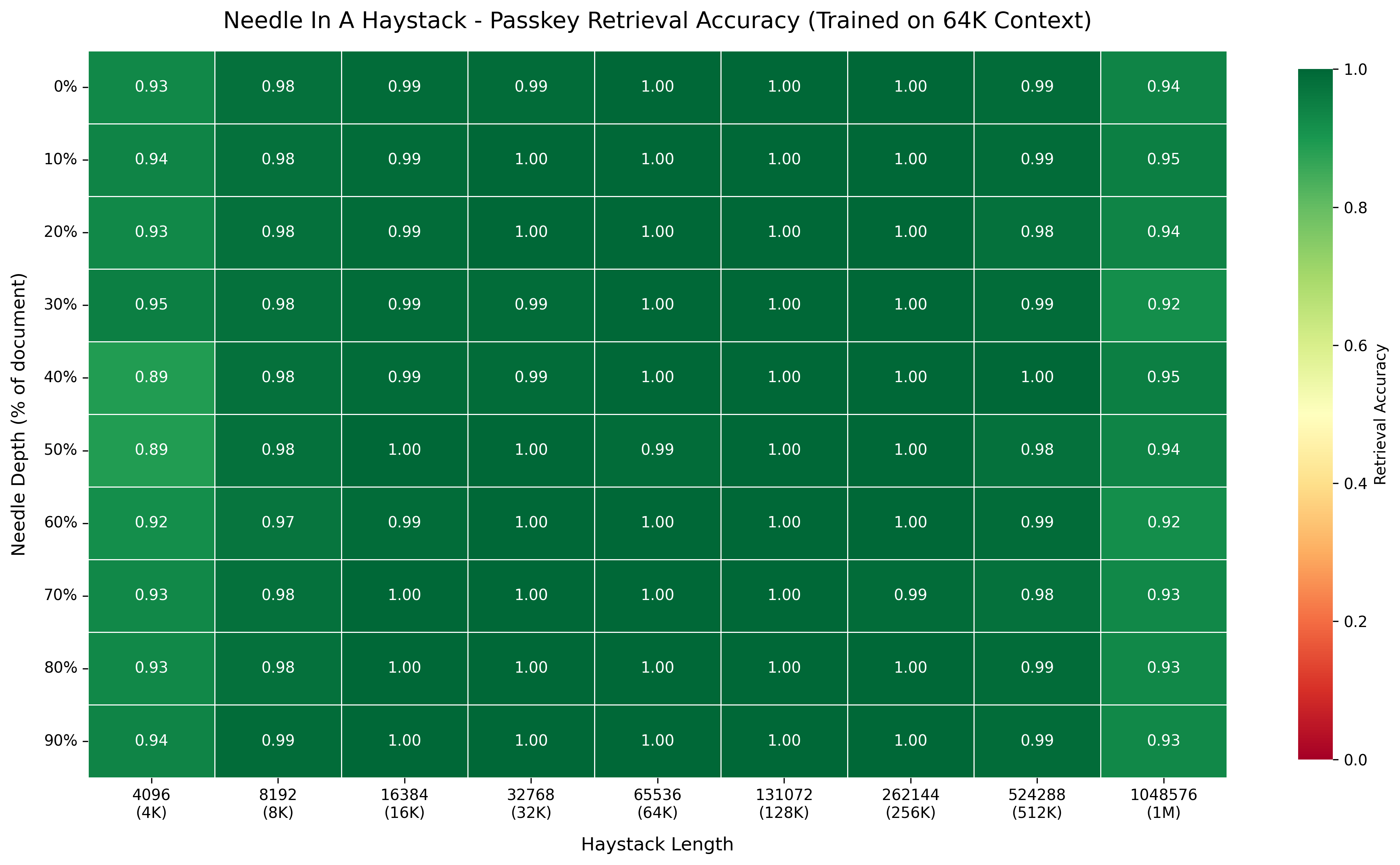}
\caption{64K Training Window}
\label{fig:niah64k}
\end{subfigure}
\caption{Multi-needle in a haystack heatmap. The horizontal axis is context length (tokens), the vertical axis is the depth interval of the needle, and the values are retrieval accuracy.}
\label{fig:niah}
\end{figure}

\paragraph{Results.}
Figure \ref{fig:niah} presents the depth-length heatmaps on $1$K--1M (1,048,576) tokens for models with 8K and 64K training windows; color and value indicate retrieval accuracy. Within their respective effective extrapolation ranges, neither model exhibits obvious "lost in the middle" phenomena, with accuracy varying smoothly across depth intervals. The 8K trained model maintains over $94\%$ accuracy when extrapolated to 256K; the 64K trained model maintains $92\%$--$95\%$ accuracy across depth intervals when extrapolated to 1,048,576 tokens (approx. $16\times$ the training length). It should be noted that these results come from smaller scale ($d=512$, 10 layers) validation models intended to test the architecture's extrapolability, rather than for direct comparison with larger-scale instruction models.

\subsection{VRAM and Training Speed}
Table \ref{tab:vram} compares the standard decoder-only model, Base with full history attention, and ProxyFormer under 16\,GB VRAM and batch size 1. Both standard decoder-only and Base require about 15.6\,GB to train approx. 20K lengths; ProxyFormer requires only 2.9\,GB for the same 20,992 token history, and training speed increases from 1.3\,it/s to 16.10\,it/s. When history length is increased to 716,800 (0.7M), ProxyFormer VRAM is 15.1\,GB, still trainable on the same 16\,GB GPU, whereas Base cannot run at that length. This comparison directly verifies the complexity analysis in Section \ref{sec:complexity}.

\begin{table}[H]
\centering
\small
\caption{VRAM and training speed comparison under a 16\,GB GPU with batch size 1.}
\label{tab:vram}
\begin{tabular}{lrrrr}
\toprule
Model & History len. & Gen. len. & VRAM (GB) & Thr. (it/s) \\
\midrule
Standard decoder-only & 0 & 20,496 & 15.6 & 2.6 \\
Base (full-history attention) & 20,992 & 128 & 15.6 & 1.3 \\
ProxyFormer ($P=64$) & 20,992 & 128 & \textbf{2.9} & \textbf{16.10} \\
ProxyFormer ($P=64$) & 716,800 & 128 & 15.1 & 2.70 \\
\bottomrule
\end{tabular}
\end{table}

\subsection{Image Generation}
Table \ref{tab:image} gives the FID and Inception Score for image generation. ProxyFormer-JIT obtains FID 16.21 and IS 8.42 on CIFAR-10, and FID 2.06 and IS 1.95 on MNIST; JLT obtains FID 4.36 and IS 1.93 on MNIST. Compared to current large-scale diffusion models, these scores are not SOTA, but they demonstrate that the same proxy compression and top-down reconstruction mechanism can be applied to both pixel-space and latent-space flow matching paradigms.

\begin{table}[H]
\centering
\caption{Image Generation Evaluation (Lower FID is better, higher IS is better).}
\label{tab:image}
\begin{tabular}{lccccc}
\toprule
Dataset & Method & Space & FID $\downarrow$ & IS $\uparrow$ \\
\midrule
CIFAR-10 & JIT & Pixel & 16.21 & \textbf{8.42} \\
MNIST & JIT & Pixel & \textbf{2.06} & 1.95 \\
MNIST & JLT & Latent & 4.36 & 1.93 \\
\bottomrule
\end{tabular}
\end{table}

\section{Discussion and Limitations}

\subsection{Source of Advantages}
Our preliminary results support the following four mechanistic assessments. First, \emph{proxy space interaction} reduces the cost of long-distance token communication from $L^2$ to $(L/P)^2$, which is the main source of VRAM and speed gains. Second, the \emph{local residual stream} provides recoverability for compression: even if a layer's compression is lossy, deeper layers can still access fine-grained information, which is likely a reason for the stable accuracy across depth in NIAH. Third, the \emph{proxy KV Cache} allows history information to exist in an extremely short state during inference, obtaining linear caching without eviction-based compression. Fourth, \emph{implementation simplicity}: compression, decompression, and proxy interaction are all composed of standard operators such as reshape, linear projection, convolution, and standard attention, facilitating replication, modification, and deployment in mainstream frameworks.

\subsection{Limitations}
\begin{itemize}[leftmargin=2em]
    \item \textbf{Limited benchmark coverage}: Current public results focus on WikiText PPL and multi-needle NIAH, and have not yet been evaluated at scale on benchmarks like LongBench, RULER, MMLU, or general dialogue;
    \item \textbf{Small model scale}: The experimental models are validation scale at $d_{\mathrm{model}}=512$, 10 layers; scaling laws and training stability on larger models have yet to be verified;
    \item \textbf{Compression module choices}: The main results reported in this paper use fixed partitioning and linear/reshape compression; ablations on variants such as dynamic routing, overlapping patches, attention compression, and cross-layer heterogeneous dimensions are yet to be completed;
    \item \textbf{Limited visual task scale}: Image experiments are limited to MNIST and CIFAR-10, and have not been systematically compared with DiT/LDM-class models on high-resolution text-to-image generation;
    \item \textbf{Theoretical analysis}: The impact of the proxy stream on the expressive capacity of function classes and the upper bounds of compression error propagation with respect to depth and compression ratio require further formalization.
\end{itemize}

\subsection{Future Work}
Next steps include: verifying ProxyFormer's scaling laws on 1B--7B scale models; studying dynamic routing partitioning based on saliency and semantic clustering; and further integrating proxy-vector interaction with efficient sequence modeling advances.

Specifically, we plan to migrate Mamba, SSM, and other linear-time sequence models \cite{gu_mamba_2024,dao_transformers_2024}, as well as the fixed-length or constant-time state update strategies of TLinFormer and TConstFormer \cite{tang_rethinking_2025,tang_tlinformer_2025}, onto the proxy-vector sequence. The goal is to gradually replace the current proxy attention with linear- or near-constant-cost proxy state evolution, thereby further reducing proxy-level computation and KV cache. In particular, the $O(1)$-cache idea of TConstFormer is naturally complementary to ProxyFormer's block compression; their combination may lead to a hierarchical state management scheme of ``proxy compression + constant cache'' for ultra-long streaming decoding.

Furthermore, we plan to extend the same architecture to higher-dimensional data such as 3D point clouds, voxels, and videos, evaluate JLT in high-resolution text-to-image and video generation, and conduct controlled comparisons with StreamingLLM, H$_2$O, Infini-attention, and Landmark Attention on standard long-context benchmarks.

\section{Conclusion}

This paper proposed ProxyFormer, a dual-stream architecture centered on proxy vectors. By establishing a closed loop between the local stream and proxy stream—comprising bottom-up compression, global interaction in proxy space, and top-down decompression and injection—ProxyFormer reduces the computation and caching costs of global interaction for ultra-long sequences by approximately $P^2$ and $P$ times, respectively, without discarding the local residual stream. This architecture can uniformly describe 1D sequences, 2D images, 3D point clouds/voxels/videos, and higher-dimensional tensors, and can be simply implemented with common operators like reshape, linear projection, convolution, and standard attention. Multi-level factorization further controls the parameter size of compression modules at high compression ratios, a reinvestment mechanism on the channel dimension provides potential for further increasing compression ratios, and the proxy KV cache significantly reduces the memory and latency of long-history inference. Preliminary experiments have verified the architecture's feasibility in language model perplexity, 1,048,576-token multi-needle in a haystack, trainable length under 16\,GB VRAM, and pixel/latent-space flow matching image generation. The authors hope this architecture provides a new fundamental component for low-cost modeling of long contexts and high-resolution generation.

\bibliographystyle{unsrt}
\bibliography{references}

\clearpage
\appendix
\section{Implementation and Reproduction Details}

\subsection{Code and Reproducibility}
The complete source code and training/evaluation configurations are publicly available at:

\begin{center}
\url{https://github.com/simonFelix-Ai/proxy-former.git}
\end{center}

\subsection{Intellectual Property and Licensing}
The core ProxyFormer architecture described in this paper has been submitted for patent protection and is currently patent pending. Making the source code public does not constitute an implied license for commercial implementation of the architecture. Academic and non-commercial research use follows the license in the repository; any commercial deployment, for-profit corporate research and development, or product integration must be separately negotiated with and authorized by the author.

\end{document}